\pdfoutput=1

\documentclass[11pt]{article}
\usepackage{graphicx}

\usepackage{acl}

\usepackage{times}
\usepackage{latexsym}

\usepackage[T1]{fontenc}

\usepackage[utf8]{inputenc}

\usepackage{microtype}

\usepackage{inconsolata}

\usepackage[compact]{titlesec}
\usepackage{xspace}
\usepackage{amsmath}

\newcommand{\rev}[1]{\textcolor{black}{#1}}
\newcommand{\rerev}[1]{\textcolor{black}{#1}}

\newcommand{\crshort}[1]{\textcolor{black}{#1}}
\newcommand{\crrev}[1]{\textcolor{black}{#1}}
\newcommand{\nbcrrev}[1]{\textcolor{black}{#1}}

\title{Large \textit{Human} Language Models: A Need and the Challenges}

\author{Nikita Soni$^1$,  H. Andrew Schwartz$^1$,  {\bf João Sedoc$^2$,  Niranjan Balasubramanian$^1$} \\
$^1$Stony Brook University,  $^2$New York University \\
\texttt{\{nisoni, has, niranjan\}@cs.stonybrook.edu}, \texttt{jsedoc@stern.nyu.edu}}

\begin{document}
\maketitle
\begin{abstract}
As research in human-centered NLP advances, there is a growing recognition of the importance of incorporating human and social factors into NLP models. At the same time, our NLP systems have become heavily reliant on LLMs, most of which do not model authors. To build NLP systems that can truly understand human language, we must better integrate human contexts into LLMs. This brings to the fore a range of design considerations and challenges in terms of what human aspects to capture, how to represent them, and what modeling strategies to pursue. To address these, we advocate for three positions toward creating large \textit{human} language models (LHLMs) using concepts from psychological and behavioral sciences: First, LM training should include the human context. Second, LHLMs should recognize that people are more than their group(s). Third, LHLMs should be able to account for the dynamic and temporally-dependent nature of the human context. We refer to relevant advances and present open challenges that need to be addressed \rev{and their possible solutions} in realizing these goals.
\end{abstract}

\section{Introduction}
\begin{figure}[!t]
    \centering
    \includegraphics[width=0.40\textwidth]{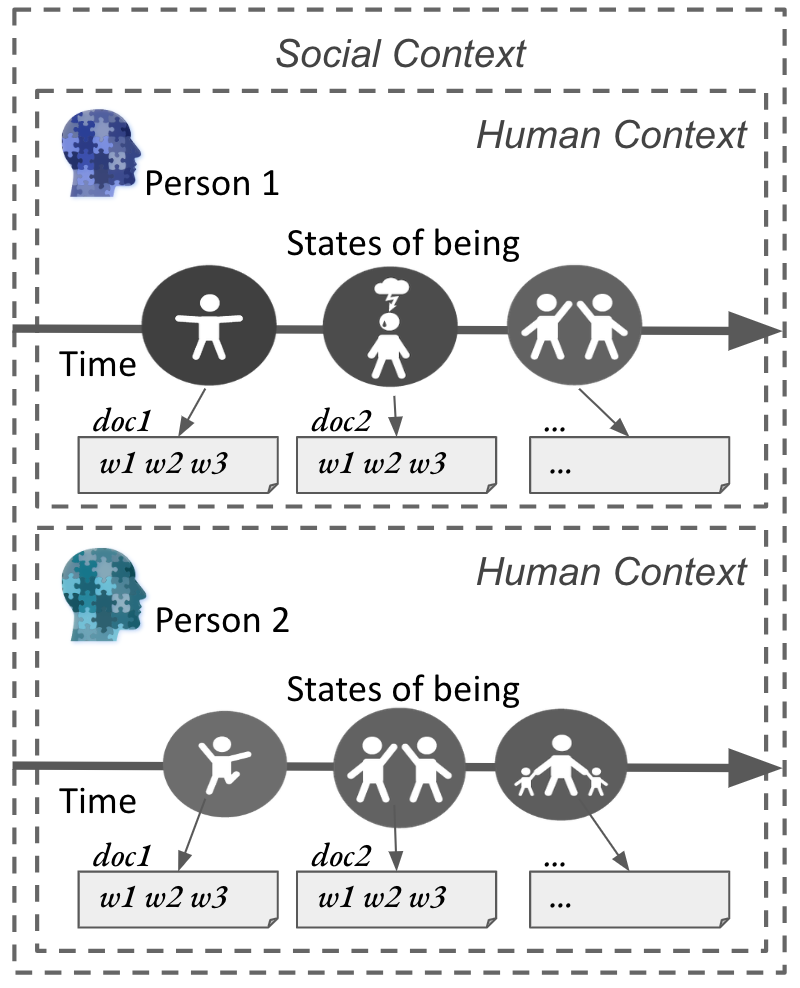}
    \caption{Language expresses the changing human states of being over time. To truly understand human language, language models should have the advantage of the \textit{dynamic human context} along with the context of its neighboring words.}
    \label{fig:human_lang_and_states}
\end{figure}

Language is a fundamental form of \textit{human} expression and communication of thoughts, emotions, and experiences. Learning the meaning of words extends beyond syntax, semantics, and the neighboring words. To truly understand human language, we must look at words in the context of the human generating the language. Figure \ref{fig:human_lang_and_states} depicts a view of how our language is moderated by our somewhat stable and changing human states of being over time \citep{fleeson2001toward, mehl2003sounds, heller2007dynamics}. 

Progress in human-centered NLP research has established the importance of modeling human and social factors, presenting a compelling argument that learning language from linguistic signals alone is not adequate \citep{hovy_social_2018, bisk_experience_2020, flek_returning_2020}, and noting that feelings, knowledge and mental states of the speaker and listener referred to as the ``Theory of Mind'' \citep{Flavell2004TheoryofMindDR}, along with other social context variables are vital to language understanding \citep{bisk_experience_2020, hovy_importance_2021}. This need is backed by a wealth of empirical evidence demonstrating the benefits of modeling human and social factors \citep{volkova_exploring_2013, hu_exploiting_2013, bamman_contextualized_2015, lynn_human_2017, radfar_characterizing_2020}, and personalized models \citep{delasalles_learning_2019, jaech-ostendorf-2018-personalized, king-cook-2020-evaluating, welch_exploring_2020}.

In parallel, with the advent of Transformers \cite{vaswani2017attention}, there have been many advances in language modeling \citep{devlin-etal-2019-bert, dai-etal-2019-transformer, liu2019roberta, radford2019language} yielding Transformer-based large language models (LLMs) as the base of most current NLP systems. LLMs train on a pre-training task and are capable of being applied to a broad set of NLP tasks producing state-of-the-art results. However, these language models create word representation 
\crrev{without }explicitly account\crrev{ing} for the context of the authors. 

Moreover, a person's language can be considered in the rich and complex human context that spans a 
\crrev{multitude} of aspects.
\begin{quote}
    \textit{[S]peakers design their utterances to be understood against the common ground they share with their addressees—their common experience, expertise, dialect, and culture.} -
\citet{Schober1989UnderstandingBA}
\end{quote}

\begin{figure}
    \centering
    \includegraphics[width=0.5\textwidth]{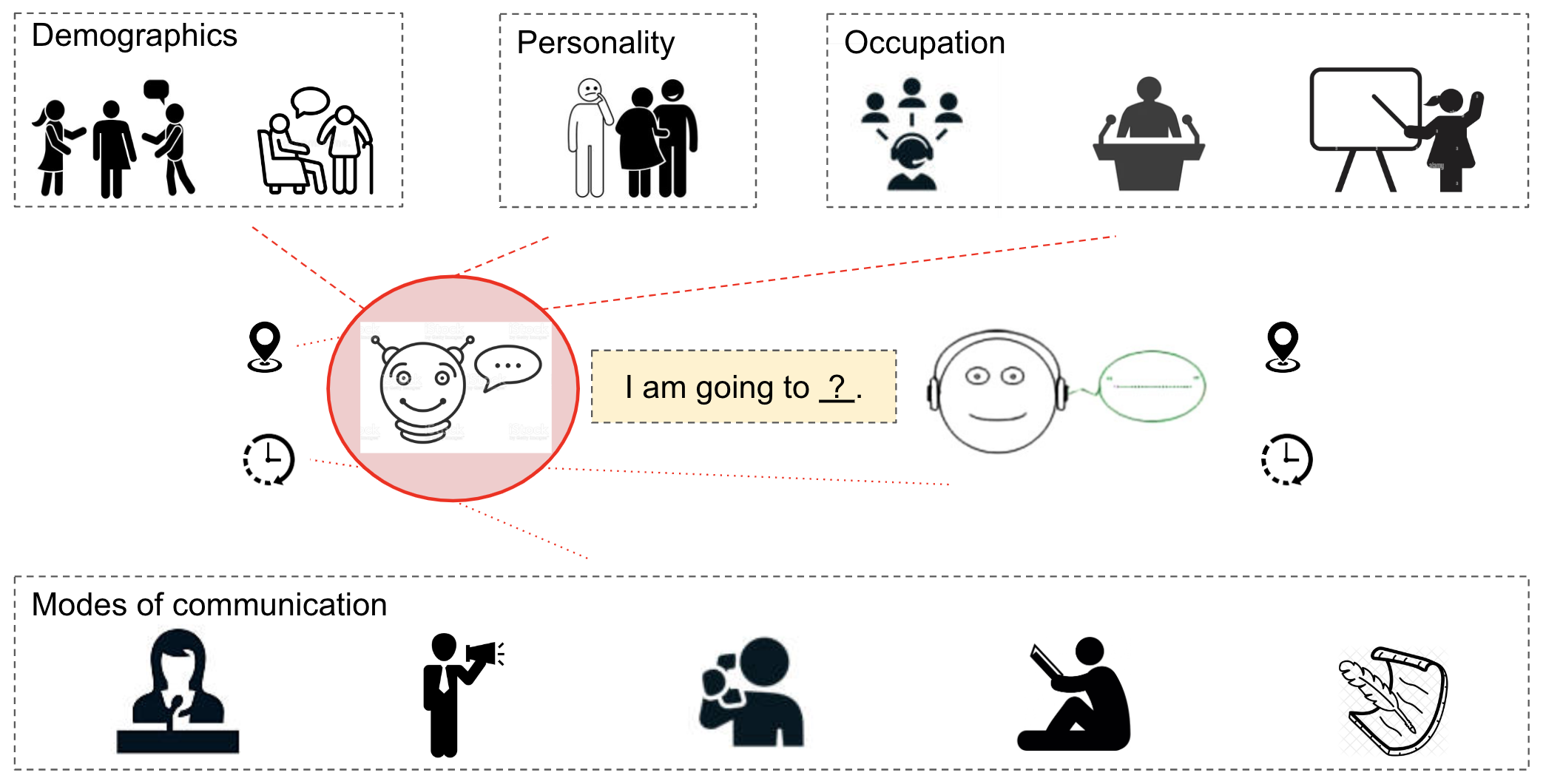}
    \caption{Language is moderated by multiple factors like \textit{who} is speaking to \textit{whom}, \textit{where}, \textit{when}, and other factors like demographics, personality, occupation, modes of communication, etc. The author's language is highly dependent on their context, which is referred to as their \textit{human context}.}
    \label{fig:human_context_example}
\end{figure}

Figure \ref{fig:human_context_example} illustrates an extensive set of factors that can be considered ``human context'' which affects how one generates language. 
A sentence that begins with the phrase ``I'm going to...'', can be continued in various ways depending on several factors such as (a) \textit{who} is speaking, (b) \textit{where} are they / in what situation and (c) \textit{when} are they speaking, and (d) to \textit{whom} the sentence is addressed including their own time and place. 
Specific examples of factors include age, personality, occupation, etc., and the forms and modes of communication like public speaking, letter writing, books, phone conversations, etc. 
The speaker's language is, thus, highly dependent on the speaker's states, traits, social and environmental factors~\citep{boyd2021natural}, which, collectively, are referred to as the \textit{human context}.

\nbcrrev{LLMs can benefit immensely from integrating the human context to truly understand human language but this entails multiple challenges. 
LLMs can be seen as containing a multitude of personas, and when prompted or primed appropriately can assume a specific one~\cite{patel2022low}. \crrev{Recently, such user-centric prompting has been employed for personalized recommender systems~\citep{doddapaneni-etal-2023-towards}, dialog systems \citep{gao2023chat}, and measuring political biases and fairness \citep{feng-etal-2023-pretraining}.} Models such as GPT-3 and ChatGPT demonstrate potential for simulating some forms of human context, especially in generative tasks~\cite{reif-etal-2022-recipe}. Continued scaling --- building bigger models trained on larger amounts of data --- will continue to improve these abilities.
However, there are two fundamental limitations to this paradigm. First, models do not explicitly handle the multi-level structure (documents connected to people) necessary for modeling the richness of human context. Second, newer paradigms of in-context learning and user-centric prompting can benefit specific settings such as personalization but are still limiting LLMs from making full use of the human context more broadly. Recent evaluations and benchmarks reveal that prompting is insufficient to capture the richness of the human and social context \citep{salemi2023lamp, choi-etal-2023-llms}.} 

Instead, in this work we call for a more direct and explicit integration of the human contexts when building language models. In particular, we advocate for including the human context directly in language model training, building rich human contexts that account for the fact that people are more than their groups, and the dynamic and temporal-dependent changes to their states of being.
In short, we call for building large \textit{human} language models as a step towards better understanding the human language. \crrev{We motivate our positions using insights from a large body of past work and discuss shortcomings throughout the text.} \rev{Furthermore, we discuss open challenges in realizing this vision and their possible solutions.}

\crrev{Social context \citep{hovy_importance_2021} encompasses the human context but is not limited to it.} \crrev{In this work, we focus our vision of LHLMs in the scope of human context.}
\nbcrrev{Human context goes far beyond what can be captured from text modality alone. For example, other modalities such as gestures, speech, and body language are also a significant part of humans and their thought processes which gives us a more holistic picture of understanding human expression.} \crrev{However, in this work, we
\crrev{limit ourselves to the}
human context derived from language. We discuss limitations in detail in Section \ref{limitations}. Furthermore, LHLMs by their very nature are associated with sensitive user information and have the potential to be misused. Thus, it becomes essential to adopt a responsible release strategy for such models.} \nbcrrev{We discuss a range of ethical considerations and privacy concerns and implications in Section \ref{ethical}.}

\section{Position 1: LM training should include the human context.}
\subsection{Motivation}
\crrev{\citet{2bca0cca-e964-3217-ab80-6209b4628c61}} describe\crrev{s} a fallacy in statistical models of the world pertaining to modeling individual observations that are part of a group, as if they are independent, a so-called \textit{ecological fallacy}\crrev{.}
Current LLMs exhibit a form of this ecological fallacy, whereby text sequences \crrev{written by an}
author are treated as if independent and miss the opportunity to capture dependence~\cite{soni-etal-2022-human}. \crrev{Conversely, in the absence of a notion of authors, the LLMs can be seen as modeling the documents from many different people as if they were generated by a single universal author.}

Motivated by this need for interpreting language in its human context and inducing inter-dependence between different text sequences from an individual, we posit the need to train our base large language models with the human context. 
One broad way to frame human context-aware language modeling is as follows:

\begin{equation*}
Pr(\mathbf{X|H}) = \prod_{i=1}^n Pr(x_{i} |X_{1:i-1}, H).
\end{equation*}

This \textit{human language modeling} problem generalizes the regular language modeling problem of predicting the next word conditioned on the previous words in a text sequence $X$ to also condition on a human context $H$\footnote{This framing can also be formulated for non-autoregressive language modeling.}. To train LLMs for this human language modeling problem, we need methods to both represent the human context and to include it in our training objective.

\subsection{Past Work}
\rerev{A rich body of prior work sought to include human contexts in NLP models, broadly falling into two categories:} ones that are closer to the human language modeling frame, and others that are post hoc adaptations of models with human contexts.

\paragraph{Human context-aware language models.}
Some work on personalized language models account for human contexts through user embeddings 
\rerev{and show improvements in predicting mental health like depression \citep{wu2020author2vec}, and user attributes like demographics \citep{benton-etal-2016-learning}, 
and occupation \citep{li2015learning}}. \crshort{These focus more on creating user representations and less on informing language models with the human context.} Others pursue continued training on the language of specific users to build user-specific language models~\cite{wen2013recurrent, king-cook-2020-evaluating} achieving substantial gains in perplexity. \crshort{While these support the call for integrating human contexts in language modeling, we need to go beyond these user-specific models. Learning and storing separate models for each user presents a scalability challenge, as well as limit\crrev{s} the sharing of knowledge across different users thus limiting generalization.}

\citet{delasalles_learning_2019} 
\crrev{approach }a more generalized human language model \rerev{which improves perplexity by 10 points on the New York Times and Semantic Scholar corpus}. \rerev{They} 
\crrev{additionally condition on} a dynamic \crrev{learned} latent representation of the author to capture the human context \crshort{using an LSTM based architecture\crrev{. However, this learned vector does not capture } the richness of the human context \crrev{representative of the human characteristics and traits in }
the author's language \crrev{\citep{fleeson2001toward}}. Also, the model parameters seem to depend on the number of authors for the static component of the user representation. While this is better than approaches that create one model per user, the growth in parameters limits scalability and generalization.}
\citet{soni-etal-2022-human} go further towards modeling the rich dynamic human context from the author's historical language and including it in the continued training of a modified GPT-2 based model. \rerev{They use social media datasets and show LM improvements with perplexity gains of up to 20 points, and improved downstream task performance on four different tasks including sentiment analysis, stance detection, assessing personality, and estimating age.} \crrev{This provides one \crrev{direction of research} 
to scalable and generalizable LHLMs but limits the amount of historical language that can be used. }

\paragraph{Post hoc human contextualized models.}
\rerev{Two broad groups of methods} use human contexts in a post hoc fashion: Personalized application-focused models, and \crrev{d}ebiasing methods using semantic subspaces. Some examples of the first group include methods that create user-specific feature vectors \citep{jaech-ostendorf-2018-personalized, seyler_leveraging_2020} or prefixed static user identifiers \citep{mireshghallah2022useridentifier} or prefixed learned user-specific vectors \citep{zhong_useradapter_2021, li-etal-2021-personalized} to the word vectors \rerev{and show improved accuracies} for personalized sentiment analysis, personalized search query auto-completion, or personalized explainable recommendation. Others developed hierarchical modeling using historical text from a user to create personalized models to improve personality detection \citep{lynn_hierarchical_2020} and stance detection \citep{matero-etal-2021-melt-message}. In the second group, several studies focused on identifying and eliminating word vector subspaces associated with a particular bias such as gender \citep{bolukbasi2016man, wang2020double, ravfogel2020null} and religion \citep{liang-etal-2020-towards}.
The broad evidence for personalization and debiasing indicates the performance and fairness benefits of modeling human contexts. \crrev{Moreover, current methods do not take into account the speaker and addressee which is essential for uncovering situational bias.}

Given the move towards large language models as the basis for NLP, we argue that if the base LLMs can be made human context aware, we can learn better and more fair language representations to begin with.

\subsection{Challenges and Possible Solutions}
\noindent\textbf{C1: Including human context.}
Training LLMs for the human language modeling problem raises a wide range of challenges.
These include deciding how to capture the human context effectively and how to incorporate it in training. \\
\noindent\textbf{PS1:} Before the advent of LLMs, human-centered NLP mainly infused human context $H$ (e.g. demographic value of an author) into a feature space $F$ either using factor additive approaches \citep{bamman_distributed_2014, bamman_contextualized_2015, kulkarni_freshman_2016, welch_compositional_2020}:
\begin{equation*}
        P = g(F + H),
\end{equation*}

\noindent
or through user factor adaptation~\citep{lynn_human_2017, huang_neural_2019}:

 \begin{equation*}
        O = {g(z(F, H))},
\end{equation*}

\noindent
where $g$ is a model trained to output predictions $P$, and $z$ represents a form of multiplicative compositional function that is used to adapt the feature space to the human context. 

We can extend these ``pre-LLM'' approaches to LLMs by viewing the hidden states or the contextual word vectors as features. The human context can thus be added directly to the contextual word vectors similar to how position embeddings get added or via composition functions that adapt the contextual word vectors conditioning on the human context. More generally, integrating human context into Transformer based LLMs brings up many challenges in terms of modeling, interaction with downstream applications, and data processing. \crrev{We discuss these next.}

\noindent\textbf{C2: Modeling decisions.} 
Architectural decisions include: which layers to modify, where do we include the human context, how to alter the self-attention mechanism if needed, and which components (query, key, value) should include the human context if needed. \\
\noindent\textbf{PS2:} For example, \citet{soni-etal-2022-human} modify the language modeling task to include the human context as a user vector, which is derived from the author's historical text. The new Transformer-based architecture modifies the self-attention computation by using the user vector in the query representation, and recurrently updates the user vector using the hidden states from a later layer. Other works \citep{zhong_useradapter_2021, mireshghallah2022useridentifier}, as discussed earlier, simply prefix the user representation to the word embeddings when processing through the Transformer based architectures. 

The \crrev{modeling decision} questions and existing works spur us to explore many other architectural solutions for large human language models, along with suitable pre-training tasks or loss functions that include human contexts. 

\noindent\textbf{C3: Model applications.}
Another key challenge is in effectively applying the pre-trained large human language models on the target downstream tasks and applications. 

\noindent\textbf{PS3:} 
For instance, (i) the pre-training task may be built similar to downstream task training i.e., we add a classification or regression head on top of the pre-trained language model and fine-tune for target downstream tasks like a traditional large language model, (ii) the pre-trained model can be trained with downstream task-specific objective i.e., in addition to using the pre-training knowledge, we train the model parameters specific to the target downstream task objective alone, (iii) continue the pre-trained model's training in a multi-task learning setup i.e., we train for the pre-training objective as well as a downstream task-specific objective, or (iv) \crrev{explore different data processing strategies when } fine-tun\crrev{ing} the pre-trained model \crrev{for target downstream tasks}
for example, limiting the historical language context \crrev{in the fine-tuning stage.}

\noindent\textbf{C4: Data processing.} 
Processing human context from user's historical language requires \crrev{effectively handling user-level data: approaches } 
to process user-specific data which can be rather long, \crrev{and strategies to choose the right amount and relevance of the historical language to be used. First, user information adds another dimension to the data that may require creative ways of processing. Second, t}he runtime and memory complexity of the self-attention mechanism scales quadratically with the sequence length, which often limits their abilities to directly process long input sequences. \crrev{And third, answers to questions like: how much historical language is sufficient to capture the human context, whether adding more language will help build a better human context, and whether we need to process even longer documents in a single pass, among other intriguing considerations.}\\
\noindent\textbf{PS4:} Some approaches to address \crrev{these} limitation\crrev{s} include \crrev{recurrently processing all of a user's data together as a single instance \citep{soni-etal-2022-human}, and incorporating existing approaches to solve long-context processing into LHLMs. These have three broad categories:} sparsifying the attention mechanism \citep{beltagy_longformer_2020, kitaev2020reformer, qiu-etal-2020-blockwise, ye2019bp, roy2021efficient}\crrev{, }using auto-regressive recurrence-based methods \citep{sukhbaatar-etal-2019-adaptive, rae2019compressive, dai-etal-2019-transformer, yoshida_adding_2020}\crrev{, and retrieval-augmentation mechanisms~\citep{guu2020realm}. However, we still need to explore the questions regarding how much and which part of an author's historical language is sufficient to model the human context.}



\section{Position 2: LHLMs should recognize that people are more than their group(s).}
\subsection{Motivation}
Human context is not limited to a specific social and demographic group they belong to. Rather it is a mix of the multiple human attribute groups they may belong to and their unique characteristics and idiosyncrasies. Even with their groups, it is not always a binary association, there are varying degrees to which an individual might align with the group traits. 

Psychology and Psychopathology have a wealth of literature suggesting that people should not be put in discrete bins but instead should be placed in a dimensional structure by characterizing them as a mixture of continuous factors \citep{mccrae1989reinterpreting, ruscio2000informing, widiger2005diagnostic}. Further, grouping people into discrete bins often uses arbitrary boundaries which may lose the meaningful distinctions in capturing the human context. 

Cross-cultural psychology research has noted the distinctions in individualism and collectivism concurring with the predictions from Hofstede's model \citep{hofstede1984culture, hofstede1984hofstede}.
\begin{quote}
    \textit{"\crrev{[P]}eople from the collectivist culture produc\crrev{[e]} significantly more group and fewer idiocentric self-descriptions than \crrev{...} people from the individualist cultures"} -\citet{bochner1994cross}
\end{quote}

These suggest that it is vital to allow for flexible interactions between individualistic and collectivist aspects of the human context.

Moreover, the rich diversity in people cannot be captured effectively by modeling a narrow sample of variation in human factored groups. In behavioral sciences, \citet{henrich2010weirdest} bring to attention that most of the research in the field is often limited to humans belonging to the WEIRD (Western, Educated, Industrialized, Rich, and Democratic) group. 
They argue that this narrow group is mostly an outlier as a representative of humanity in cross-cultural research. This provides a corresponding lesson for NLP research. We should not limit ourselves to a narrow spectrum of specific human factors \crrev{by} 
only modeling outliers in the human context. 

Motivated by these ideas from psychology and behavioral sciences, we argue for breadth, depth, and richness in modeling the human context when training large human language models.

\subsection{Past Work}
A huge body of work in human-centered NLP has shown the importance of modeling human attributes like demographic factors and social context, and latent human variables in natural language processing. These include works that model factors that are either known explicitly from questionnaires, social profiles, or inferred from the user's language, with the aim of grouping people to analyze language variations among different groups.

\paragraph{Wide variety of human factors.}
There are many types of human factors that can influence a person's language.
Cross-cultural differences and demographics like gender \citep{volkova_exploring_2013} and age \citep{hovy_demographic_2015} have been shown to influence the perceived meaning of words and aid in multiple text classification tasks \citep{huang_neural_2019}, and machine translation \citep{mirkin_motivating_2015, rabinovich_personalized_2017}. Several studies have also exploited benefits from social relations \citep{huang-etal-2014-enriching, yang_overcoming_2017, zeng_socialized_2017, del_tredici_you_2019} in sentiment analysis \citep{hu_exploiting_2013} and toxic language detection \citep{radfar_characterizing_2020}. Existing literature has shown correlations in language variation with personality \citep{schwartz_personality_2013}, occupation \citep{preotiuc-pietro_role_2015}, and geographical region \citep{bamman_distributed_2014, kulkarni_freshman_2016, garimella_demographic-aware_2017} illustrating distinctions in style and perspectives among different groups of people.

\paragraph{Intersectionality of human factors.}
A person's language is mediated not just by an individual factor but by the intersection of many factors. Some works \citep{bamman_contextualized_2015, lynn_tweet_2019, huang_neural_2019} have explored using multiple human factors together in their studies. Some classification tasks from different domains \citep{huang_neural_2019} have shown greater benefits in a multi-factored approach of combining gender, age, country, and region, while tasks like sarcasm detection \citep{bamman_contextualized_2015} and stance detection \citep{lynn_tweet_2019} have performed better by specific author features. \crrev{\citet{soni2024comparing} find pre-training with individual traits and group attributes help user-level tasks like assessing personality, while incorporating only the individual human context in pre-training benefits document-level tasks like stance detection.} These empirical studies indicate the need to explore different combinations of human factors for respective downstream tasks and applications.

\paragraph{Continuous representation of human factors.} 
A discrete group often relies on arbitrary boundaries and a person may belong to multiple groups in varying degrees. Thus, using a continuous representation of human factors may allow us to move away from \textit{hard memberships} in arbitrary groups to a more realistic \textit{soft membership} along factor dimensions. Prior work has illustrated language differences based on social network clusters with strong gender orientation, treating gender as more than a binary variable \citep{bamman2014gender}, or by continuous adaptation of real-valued human factors like continuous age, gender, and Big Five personality traits \citep{lynn_human_2017}.

\paragraph{Latent human factors.}
A person's language has characteristics that go well beyond those of a specific set of groups they may belong to. To capture a broader set of characteristics, some works explored deriving latent factors from a person's language \citep{wen2013recurrent, lynn_human_2017, kulkarni2018latent}. Latent linguistic factors have been shown to capture user attributes~\citep{lynn_human_2017} and differences in thoughts and emotions of people \citep{kulkarni2018latent}. Others create latent representations from user posts using bag-of-words~\citep{benton-etal-2016-learning}, sparse-encoded BERT contextual embeddings~\citep{wu2020author2vec}, and averaged GRU embeddings~\citep{lynn_hierarchical_2020}. Another approach focuses on learning embeddings, i.e., a trainable set of parameters, as latent representations of users~\citep{li2015learning, amir-etal-2016-modelling, zeng_socialized_2017, jaech-ostendorf-2018-personalized, welch_exploring_2020}.
These latent user representations and learned embeddings \crrev{yield}
benefits in multiple downstream tasks and applications.

Modeling the human context in terms of \crrev{the} groups that people belong to has pioneered advances in human-centered NLP. However, humans are more than the \crrev{discrete} groups they belong to. 
To go further, we need a representation that
recognizes the variety, and intersectionality of human factors
across continuous dimensions, as well as their unique individual characteristics.

\subsection{Challenges and Possible Solutions}
\label{privacy}
\rev{\textbf{C1: Modeling data and representational disparities.}
To capture the rich human context, we need access to datasets that provide relevant information covering users who are representative of the broad and diverse population \citep{henrich2010weirdest, johnson2022ghost}. Specifically, the challenges lie in obtaining datasets:
(1) that provide access to user identifiers and historical language \crrev{which}
allow us to differentiate the human source of the language, \crrev{and} associate explicit human attributes such as sociodemographic or personality attributes,
(2) that do not amplify representational disparities \citep{shah2020predictive} and span multiple domains such as healthcare~\citep{bean2023hospital}, customer service~\citep{adam2021ai}, and education~\citep{klein2019learning}. 
}

\noindent\rev{\textbf{PS1:} There are multiple avenues for addressing the challenges above. First, there is a wide-variety of large scale datasets that contain author Ids as metadata. For example, Amazon reviews, Reddit posts and comments, blogs, books, and news \crrev{articles,}
which can be used to train LHLMs. Second, some representational disparities can be addressed by benchmarking and balancing the types of disparities. For example, we can use various text-based human attribute inference methods to detect and balance for attributes such as age, gender, and other demographics \citep{tadesse2018personality, wang2019demographic}. Similarly, we can address cultural disparities by making use of research efforts to probe \citep{arora-etal-2023-probing}, identify \citep{gutierrez-etal-2016-detecting, lin-etal-2018-mining} and benchmark \citep{yin2022geomlama} cross-cultural differences. Third, we can also use modeling strategies that are better equipped to handle imbalanced and limited data settings. For example, there is a large body of work in low-resource settings for problems such as sentiment analysis ~\citep{priyadharshini2021overview, muhammad2023afrisenti}, hate speech detection \citep{modha2021overview}, and machine translation \citep{ranathunga2023neural}. Other notable examples include strategies for culturally grounding models using transfer learning~\citep{sun-etal-2021-cross, zhou-etal-2023-cross}, and adaptation strategies for modeling societal values~\citep{solaiman2021process}.
}

\rev{Additionally, industries with large user bases are a potential source for language data. Investing in community-wide efforts for publishing and evaluating research over proprietary data and improved industry collaborations can provide access to otherwise unavailable data which can also help further research in this area.
}

\noindent \textbf{C2: Privacy issues.}
Modeling user's personal characteristics carries the inherent risk of inadvertent privacy leaks as well as the potential for adversarial or malicious use. The challenge of guarding the privacy of individuals can be broadly categorized into 2 aspects: (1) Privacy and data control of the data subject, and (2) Licensing model usage, policies, and laws to prevent potential misuse like target marketing: As seen in the past with Cambridge Analytica Facebook dataset, a potential misuse of modeling humans is target marketing \citep{8436400, bakir2020psychological}. \\
\noindent\textbf{PS2:} \rev{\crrev{Some e}xisting laws 
aim to protect user privacy and security, such as requiring data anonymization and/or asking for consent to share data}. \crrev{For example, t}he EU General Data Protection Regulation (GDPR) \citep{lewis-etal-2017-integrating}, 
is considered one of the strongest \crrev{laws.} 
\crrev{The Italian Data Protection authority} banned 
the widespread ChatGPT services \citep{bertuzzi2023, satariano2023} citing concerns over \crrev{privacy violations and} breach\crrev{ing}
\crrev{the} EU \crrev{GDPR.}
\crrev{T}he Institutional Review Board (IRB) approvals process \crrev{is followed in the US} to protect human subjects research with most standards rooted in ethical standards involved in medical research \citep{goodyear2007declaration, miracle2016belmont}
such as protecting the rights of all research subjects or participants in terms of respect, beneficence, justice, the right to make informed decisions, and recognition of vulnerable groups.  
We should be vigilant in preventing such leaks and have strict licensing and policies to safeguard malevolent uses. Human context aware models themselves can be used towards some of these goals such as recognizing target marketing and preventing its spread. A key part here is in continuing to evolve privacy laws and policies as the models evolve and investing in studies that can better inform these decisions.

\noindent\textbf{C3: Model scalability.}
Targeting human contexts that go beyond group characteristics and include 
unique individual characteristics increases the scalability requirements on the models. The key challenge is that the model has to simultaneously capture user-specific contexts as well as scale to multiple users without corresponding increases in model parameters or creating a new model itself for each user.
Past work on personalized models have been limited by this scalability issue, whereby either models are user-specific or do not scale well. In some, a separate model is created for each user \citep{king-cook-2020-evaluating}, while in others a different user identifier is used for each user \citep{li-etal-2021-personalized, zhong_useradapter_2021, mireshghallah2022useridentifier}. \\
\noindent\textbf{PS3:} Some use a post hoc fix which handles any new user seen after training by updating the user embeddings with the new user directly during evaluation~\citep{jaech-ostendorf-2018-personalized}.
\citet{delasalles_learning_2019} adopt an LSTM-based approach with a dynamic author representation which consists of user-specific static and dynamic components. 
These approaches that learn user-specific vectors are relatively more scalable than the ones that learn user-specific models. 
\citet{soni-etal-2022-human} eliminate this dependence on user-specific vectors using a single Transformer-based model, where a recurrent user states module is trained to use authors' historical language. While this improves scalability, it is still limited in the amount of historical language it can use due to the compute requirements and context-length considerations. These ideas pave the way for further explorations of solutions to this challenge of building scalable large human language models.

\section{Position 3: LHLMs should account for the dynamic and temporally-dependent nature of human context.}

\subsection{Motivation}

\begin{quote}
    \textit{"[People] are
    embedded within time, \crrev{...} time is fundamentally important to life as it is lived, and \crrev{...} personality processes take place over time."} -\citet{larsen1989process}
\end{quote}

A person's static and dynamic human states are intertwined, where static traits influence the likelihood of entering various dynamic states across time~\citep{deyoung2015cybernetic}.
Correspondingly, a person's language expresses the changing human states and evolving emotions over time \citep{fleeson2001toward, mehl2003sounds, heller2007dynamics}. 
For the human context to be effective, it must not only be able to model the static human traits and attributes but also the more dynamic human states of being.

Temporal rhythms (e.g. diurnal and seasonal) are also known to affect human mood and behavior, which in turn manifests in their language~\citep{golder2011diurnal}. We need mechanisms that can capture the patterns of regularity or change in human language and human behavior over time. For example, studies on NLP for mental health also point to the importance of tracking moments of change over time for assessing suicidal risk~\cite{tsakalidis2022overview}.

Motivated by these ideas of changing human states and the impact of temporal aspects on human behavior and language, we posit the need for a dynamic and temporally-dependent human context.

\subsection{Past Work}

Studies that explore the dynamic nature of human context fall into two broad categories, those that: (1) dynamically update user representations to capture changing human states, and those that (2) contextualize using temporally ordered texts and other aspects that demonstrate the recurrent changes from seasonality or other cyclic patterns.

\paragraph{Recurrently updated user representations.}
As discussed earlier, recurrence mechanisms have been used for building user representations~\citep{delasalles_learning_2019, soni-etal-2022-human}. 
\crrev{It }is motivated by the need to capture author-specific features that do not change with time\crrev{, and the} 
author's \crrev{human states, topic evolution, and altering expressions that change over time.}
\citeauthor{delasalles_learning_2019} learned a dynamic latent vector using an LSTM model for this purpose, \crrev{and} \citeauthor{soni-etal-2022-human} \crrev{go} further \crrev{to} use the target user's historical texts to recurrently update the user representation. When learnt over temporally ordered language, these \crrev{methods} enable capturing the changing human states and temporal aspects as exhibited through their language. \crrev{But, these methods are limited by either the amount and specific parts of the author's historical data used or by the complete absence of it.}

\paragraph{Temporal Modeling.}
The changing human states over time highlights the need to consider the \textit{temporal} aspect of the human context and its expressions in language. 
Considering temporally ordered texts allows capturing some notion of temporality in an implicit fashion. \citet{matero-etal-2021-melt-message} introduced a missing message prediction task over a sequence of temporally ordered social media posts of the target user to build a personalized language model that helps in stance detection. \citet{tsakalidis2022overview} proposed a shared task to capture drastic and gradual moments of change in an individual's mood based on their language on social media and to identify how this change helps assess suicidal risk \citep{boinepelli-etal-2022-towards, v-ganesan-etal-2022-wwbp}. \citet{zhou2020temporal} use other temporal aspects like typical periodicity or cyclical nature, frequency, and duration to induce common sense in language models but over generic newswire texts with no direct relation to the human contexts of the authors.

We propose using recurring patterns or anomalies can better inform our dynamic human context to capture a better representation of a person as a whole. This enriched human context capturing the periodicity or anomalies in human behavior and their language can also help in multiple mental health applications and early detection.

\subsection{Challenges and Possible Solutions}

\noindent\textbf{C1: Modeling data.}
To model the dynamic and temporal changes in language, we need time information in our datasets. Assessment over time can be thought as an additional dimension to the dataset, resulting in a three-dimensional dataset \citep{larsen1989process} with user information, text, and time. While it may be possible to obtain a reasonable history of a user's language, obtaining adequate samples across all timestamps is difficult. \\
\noindent\textbf{PS1:} Thus, datasets are likely to have larger ``gaps'' in the time dimension and models may need to learn to fill or otherwise adequately handle these gaps in temporal text sequences~\citep{matero-etal-2021-melt-message}.

\noindent\textbf{C2: Modeling temporal language and temporal aspects.}
The positional encoding in a temporally ordered sequence can allow a language model to learn some temporal aspects (e.g. before/after relationships). However, more complex recurrent dynamics at different time scales (e.g. diurnal, weekly, and seasonal) may need other mechanisms that allow the model to explicitly consider the time associated with each text. This raises new challenges in encoding such time information into a temporal embedding and in getting models to use this encoded information. Last, pushing models to consider temporal information may also require developing new language modeling objectives. \\
\noindent\textbf{PS2:} Predicting what follows can often be modeled by focusing on the immediate local dependencies (in a Markovian sense). However, to force models to consider different temporal scales we can consider objectives that frame predicting what will be said after a specific temporal interval (e.g. the next day, the same day next week and so on).

\section{Conclusion}
Building upon the success of two parallels of NLP research: large language models and human-centered NLP, we envision large \textit{human} language models (LHLMs) as the base of future NLP systems. 
Previous positions taken in human-centered NLP advocate for modeling human and social factors \citep{hovy_social_2018,bisk_experience_2020,flek_returning_2020,shah2020predictive,hovy_importance_2021}. We go further and call for modeling a richer and dynamic human context in our future large language models. A rich human context captures the personal, social, and situational attributes of the person, and represents both static traits and dynamic human states of being. We put forward three specific positions as steps toward integrating this rich human context in language models to realize the vision of large \textit{human} language models. Our roadmap draws on motivations from multiple disciplines, prior advances in human-centered NLP, and organizes the range of challenges to be met in realizing this vision. We call for our NLP research community to take on the challenge of bringing humans, the originators of language, into our large language models.

\section{Limitations}
\label{limitations}
We elaborate on the three positions we take to create large \textit{human} language models in terms of the need, richness, and dynamic nature of the human context in the main paper. 
However, the scope of this position is fairly limited, focusing on the details of the human context, only giving social context a brief mention in so far as its relation to human context. 
Important social contexts affecting language include (1) cultural shifts/changes, (2) environmental events like natural disasters, and (3) multi-lingual settings (although most of our discussion is based on the psychological theory that transcends languages).
Similarly, we limit our discussion on the needs and challenges of the breadth of the domains of the human context. 
Finally, our discussion of privacy issues is also focused on the human context (refer section \ref{privacy}) and thus does not go into required social policies and its effects on language models. \crrev{Furthermore, we note that human context is not necessarily confined to the space of language. There is a broader notion of human context extending to multi-modality (for example, speech, gestures, body language, etc.) that gives us a more holistic understanding of human expression. We limit our paper's scope of human context to that inferred from language alone and leave envisioning LHLMs in a multi-modal view as part of future work.}

\section{Ethical Considerations}
\label{ethical}
Many of the main points of this paper are in themselves of ethical consideration. 
We thus use this section to discuss the uncovered considerations. 
Importantly, while we advocate for large \textit{human} language models and training them with a rich and dynamic human context, we also argue not every use case of LHLMs are of societal benefit.  
When developing LHLMs to better understand human language and to enable bias correction and fairness, one should also seek a responsible strategy for the release and use of user-level information which can sometimes be sensitive or private. \crrev{Additionally, models predicting author attributes and sociodemographic information can enable accounting for human language variation and have the potential to produce fairer and more inclusive results, but at the same time need to be considered with particular scrutiny. With the risk of identifying sensitive user information, such models can potentially lead to profiling and stereotyping. }
For such data, user consent and privacy protections are important. 
Otherwise, such models \crrev{also present opportunities for unintended harms, malicious exploitation, and }could be used for targeted content toward training set users without their awareness. 
While laws in some nations, such as the GDPR, outlaw such use cases, these have not become universal around the world yet. 

\section*{Acknowledgments}
This research is supported in part by the Office of the Director of National Intelligence (ODNI), Intelligence Advanced Research Projects Activity (IARPA), via the HIATUS Program contract \#2022-22072200005, and a grant from the CDC/NIOSH (U01 OH012476). The views and conclusions contained herein are those of the authors and should not be interpreted as necessarily representing the official policies, either expressed or implied, of ODNI, IARPA, any other government organization, or the U.S. Government. The U.S. Government is authorized to reproduce and distribute reprints for governmental purposes notwithstanding any copyright annotation therein.

\bibliography{anthology,custom}

\end{document}